\documentclass{article}
\usepackage{log_2022}            % for camera-ready version
\usepackage{booktabs}            % professional-quality tables
\usepackage{multirow}            % tabular cells spanning multiple rows
\usepackage{amsfonts}            % blackboard math symbols
\usepackage{graphicx}            % figures
\usepackage{duckuments}          % sample images

\usepackage{caption}
\usepackage{subcaption}
\usepackage{listings}
\usepackage[symbol]{footmisc}
\usepackage{adjustbox}

\usepackage[numbers,compress,sort]{natbib}

\usepackage[]{todonotes}
\usepackage{marginnote}

 \newcommand\blfootnote[1]{%
  \begingroup
  \renewcommand\thefootnote{}\footnote{#1}%
  \addtocounter{footnote}{-1}%
  \endgroup
}

\renewcommand{\thefootnote}{\fnsymbol{footnote}}

\newcommand{\hints}{\mathcal{H}}

\newcommand{\hintstackop}{\hints_\text{stack\_op}}

\newcommand{\push}{\texttt{push}}
\newcommand{\pop}{\texttt{pop}}
\newcommand{\noop}{\texttt{noop}}
\newcommand{\pointer}{\rho} % TODO different notation, rho looks like p but feels wrong
\newcommand{\change}[1]{#1}
\usepackage{array, etoolbox} 
\newcounter{rowno}
\preto\tabular{\setcounter{rowno}{0}}

\newcolumntype{R}[2]{%
    >{\adjustbox{angle=#1,lap=\width-(#2)}\bgroup}%
    l%
    <{\egroup}%
}
\newcommand*\rot{\multicolumn{1}{R{45}{1em}}}

\newlength\bigH
\newlength\smlH
\newcommand{\confintint}[2]{#1\raisebox{\dimexpr\bigH-\smlH}{\scriptsize $\pm$#2}}
\newcommand{\confint}[3][n]{\ifthenelse{\equal{#1}{b}}%
                     {\textbf{\confintint{#2}{#3}}}%
                     {\ifthenelse{\equal{#1}{u}}{\underline{\confintint{#2}{#3}}}{\ifthenelse{\equal{#1}{bu}}{\textbf{\underline{\confintint{#2}{#3}}}}{\confintint{#2}{#3}}}}}

\title[Recursive Algorithmic Reasoning]{Recursive Algorithmic Reasoning}

\author[J. Jürß, D. Jayalath, et al.]{%
Jonas Jürß\footnotemark[1] \hspace{1pt}\\
\institute{University of Cambridge}\\
\email{jj570@cl.cam.ac.uk}\And
Dulhan Jayalath\thanks{Equal contribution. † Work done while author was at the University of Cambridge.}  \hspace{3pt}\footnotemark[2]\\
\institute{University of Oxford}\\
\email{dulhan@robots.ox.ac.uk}\And
Petar Veličković\\
\institute{Google DeepMind}\\
\email{petarv@deepmind.com}
}

\usepackage{pifont}% http://ctan.org/pkg/pifont

\newcommand*\colourcheck[1]{%
  \expandafter\newcommand\csname #1check\endcsname{\textcolor{#1}{\ding{51}}}%
}
\newcommand*\colourcross[1]{%
  \expandafter\newcommand\csname #1cross\endcsname{\textcolor{#1}{\ding{55}}}%
}
\colourcheck{gray}
\colourcheck{black}
\colourcross{gray}
\colourcross{black}

\begin{document}

\maketitle

\begin{abstract}
Learning models that execute algorithms can enable us to address a key problem in deep learning: generalizing to out-of-distribution data. However, neural networks are currently unable to execute recursive algorithms because they do not have arbitrarily large memory to store and recall state. To address this, we (1) propose a way to augment graph neural networks (GNNs) with a stack, and (2) develop an approach for sampling intermediate algorithm trajectories that improves alignment with recursive algorithms over previous methods. The stack allows the network to learn to store and recall a portion of the state of the network at a particular time, analogous to the action of a call stack in a recursive algorithm. This augmentation permits the network to reason recursively. We empirically demonstrate that our proposals significantly improve generalization to larger input graphs over prior work on depth-first search (DFS).
\end{abstract}

\section{Introduction}
\label{sec:intro}
% \todo{Cite our own work \cite{tinypaper}. Possibly update the bibtex to include ICLR}
Mimicking classical algorithms by executing them with neural networks can bring many of the benefits of reasoning to black-box supervised learning\blfootnote{Our code is available at \url{https://github.com/DJayalath/gnn-call-stack/}.}. One of these benefits is taking advantage of the guarantees that algorithms provide. Classical algorithms are guaranteed to be correct no matter the size of their input. For example, sorting algorithms are correct regardless of the length of the input array. However, neural networks provide no such confidence---they will usually not be correct for problem instances larger than those they were trained on (failing to generalize out of distribution). If neural networks could learn to reason like algorithms, they could gain the substantial generalization properties that algorithms posses \cite{xu_what_2020, velickovic_neural_2021}.

On the flip side, the use of neural networks for reasoning can bring speed and redundancy benefits that classical algorithms are unable to provide. As first demonstrated by \citet{li_strong_2020}, neural networks that mimic algorithms can outperform hand-coded solutions, in terms of \change{number of steps, in limited problem instance sizes}. More recently, \citet{numeroso2023dual} showed that, by reasoning in continuous space, neural networks can even learn to execute algorithms when some input features are missing. Consequently, neural networks also have clear advantages over classical algorithms.

Given this dichotomy, recent work in \textit{neural algorithmic reasoning (NAR)} \cite{velickovic_neural_2021} studies how to gain the benefits of both classical algorithms and neural networks by executing algorithms using neural nets. Many of these algorithms are naturally amenable to graph representations as they can be viewed as manipulations of sets of objects and the relations between them \cite{velickovic_neural_2021, velickovic_clrs_2022}. Hence, contemporary approaches in NAR train GNNs with a recurrent state to execute algorithms by predicting the state of the algorithm across time steps \cite{xu_what_2020, velickovic_clrs_2022}. For example, given the insertion sort algorithm, an input array can be represented as a chain of connected nodes, where at each time step, a GNN is applied to evolve the node embeddings. These node embeddings can predict the algorithm's state at this step, such as the position of the insertion pointer. Thus, the GNN mimics the steps of the algorithm.

While this paradigm allows GNNs to execute a range of classical algorithms, they cannot execute \textit{recursive} algorithms (e.g., DFS) in general. This is because reasoning recursively requires a neural network to have memory at least large enough to store as many states as the maximum recursion depth of the problem. As a typical neural network is fixed in size, it cannot store all the states required to reason about an arbitrarily large recursive problem. \change{\citet{DeltangChomskyHierarchy} empirically show that only networks with structured memory (e.g., a stack) generalize on context-sensitive tasks such as DFS.}

% TODO \cite{DeltangChomskyHierarchy}
% d counter-language tasks, and only
% networks augmented with structured memory (such as a stack or memory tape) can
% successfully generalize on context-free and context-sensitive tasks

To address this fundamental issue, we propose a framework for augmenting GNNs with stack memory. Inspired by call stacks in computer programs, this augmentation enables the network to learn how to save and recall states. In addition, we identify several key improvements for sampling algorithm trajectories in the CLRS-30 algorithmic reasoning benchmark \cite{velickovic_clrs_2022}. These improvements allow the network to more closely structurally resemble a recursive algorithm.

We test our framework by implementing two methods of augmenting GNNs with a stack. We evaluate these approaches on the benchmark, empirically observing that our stack-based methods outperform standard GNNs (including the work in CLRS-30) on out-of-distribution generalization. Moreover, through a set of ablation experiments, we find support for our suggested improvements and discover additional modifications which further improve out-of-distribution generalization performance.

Our insights may be practical beyond DFS. The execution path of a recursive function's call graph is precisely a depth-first search. Therefore, DFS can in principle be used to express all other recursive algorithms given that the execution path is known upfront. Consequently, we believe that our analysis will be beneficial for algorithmic reasoning across recursive problems where this is the case. Nevertheless, we also more generally demonstrate how some of our insights can be applied to recursive algorithms that are not DFS. \change{Additionally, we note that by supporting recursive reasoning, we enable the execution of looping control flows which have previously been difficult for neural networks to learn \citep{li_strong_2020}. By instead realizing them as tail recursion, they can be executed under our architecture.}

Our main contributions are:
\begin{enumerate}
    \itemsep0em
    \item A novel neural network architecture that uses a stack to learn to save and recall state exactly; this architecture significantly outperforms previous work \citep{velickovic_clrs_2022}, while using less memory, on out-of-distribution generalization when learning DFS.
    \item Improvements to the sampling of algorithm trajectories which lead to closer alignment of GNNs in NAR with recursive reasoning.
\end{enumerate}

\section{Background}
\label{sec:background}

\subsection{The CLRS Algorithmic Reasoning Benchmark}
\label{ssec:CLRS}

Towards the goal of unified evaluation in NAR, \citet{velickovic_clrs_2022} introduced CLRS-30: a benchmark for evaluating GNNs on algorithmic tasks \textit{and} a standardized neural model for algorithmic reasoning. CLRS-30 measures NAR performance on a set of 30 curated algorithms, aligning closely to the definitions in \citet{cormen_introduction_2009}'s foundational algorithms textbook: \textit{Introduction to algorithms}.

The neural model represents the inputs and outputs of an algorithm (and the relations between them) as a graph $G = (V, E)$. At each step of training, CLRS-30 provides ground truth values for the state of variables in an algorithm---referred to as \textit{hints} \cite{velickovic_clrs_2022, bevilacqua_neural_2023}. The \textit{left} and \textit{right} pointers in the quicksort algorithm are examples of hints. The network learns to predict the state of these hints, which are encoded as vectors, at each step of computation. By learning to predict hints, the network's reasoning may align more closely with the computation of the algorithm.

% Since we learn an algorithm through multiple intermediate steps, GNNs are applied as a recurrent component. As an example, a GNN can reason like insertion sort by evolving the node embeddings of the chain (analogous to updating the elements of the input array) over some time steps, mimicking the steps of computation in the true insertion sort algorithm.

\begin{figure*}
    \centering
    \includegraphics[width=0.8\textwidth]{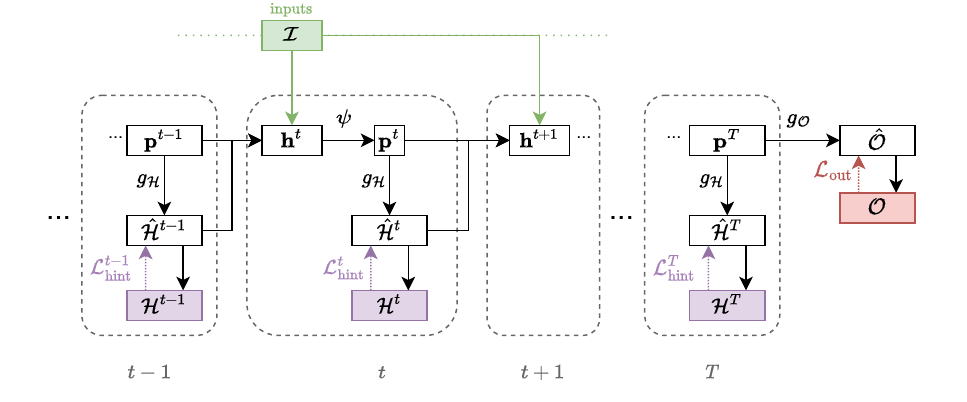}
    \caption{\textbf{Recurrent steps in CLRS-30.} The processor network takes the features $\mathbf{h}^t$ and produces the processed features $\mathbf{p}^{t}$. These are used to predict the hints $\hat{\mathcal{H}}^t$ for that step. A loss is calculated between the ground truth hints and the predicted hints. The hints can optionally be teacher forced with some probability. The predicted hints are accumulated with the processed features and algorithm inputs to form the encoded features $\mathbf{h}^{t+1}$ for the next step. In the final step, the processed features are used to predict the output $\hat{\mathcal{O}}$. A loss is calculated between this and the ground truth output of the algorithm.}
    \label{fig:clrs}
\end{figure*}

Figure \ref{fig:clrs} provides a high-level unfolded view of the recurrent steps in the CLRS-30 neural model. The inputs to one step are node inputs $\mathbf{x}_i$, edge inputs $\mathbf{e}_{ij}$, and graph inputs $\mathbf{g}$. These inputs are defined by the algorithm and hints are included as part of them (where hints can belong to nodes, edges, or the graph). They are encoded with linear layers $f$ into node, edge, and graph features $\mathbf{h}_i, \mathbf{h}_{ij}, \mathbf{h}_g \in \mathbb{R}^{d_\mathbf{h}}$ where
\begin{align}
    \mathbf{h}_i^t = f_n(\mathbf{x}_i^t) \hspace{2em} \mathbf{h}_{ij}^t = f_e(\mathbf{e}_{ij}^t) \hspace{2em} \mathbf{h}_g^t = f_g(\mathbf{g}^t),
\end{align}
and $d_\mathbf{h}$ defines the dimensionality of these features\footnote{This is the typical case. Node, edge, and graph features can also be given different dimensions.}. The features are passed through a processor network $\psi$ (a GNN) which outputs processed node and edge features $\mathbf{p}_i, \mathbf{p}_{ij} \in \mathbb{R}^{d_\mathbf{h}}$ such that
\begin{align}
\label{eq:processor}
\mathbf{p}_i^{t}, \mathbf{p}_{ij}^{t} = \psi(\mathbf{h}_i^t, \mathbf{p}_i^{t-1}, \mathbf{h}_{ij}^t, \mathbf{h}_g^t)
\end{align}
where $\mathbf{p}_i^{t-1}$ is a recurrent state carried forward from the previous time step. These processed features are decoded to predict hints
\begin{align}
\hat{\mathcal{H}}^t = g_{\mathcal{H}}(\{\mathbf{p}_i^{t}\mid i \in V\}, \{\mathbf{p}_{ij}^{t} \mid (i, j) \in E\})
\end{align}
where $g_{\mathcal{H}}$ is the \textit{hint decoder}. In the final step $T$, the neural model also predicts the output of the algorithm
\begin{align}
\hat{\mathcal{O}} = g_{\mathcal{O}}(\{\mathbf{p}_i^{T}\mid i \in V\}, \{\mathbf{p}_{ij}^{T} \mid (i, j) \in E\})
\end{align}
where $g_{\mathcal{O}}$ is the \textit{output decoder}. If this is not the final step, hint predictions are aggregated with the algorithm inputs in the following step to form the next inputs $\mathbf{x}_i^{t+1}$, $\mathbf{e}_{ij}^{t+1}$, and $\mathbf{g}^{t+1}$. In training, hints can optionally be teacher forced (i.e., predicted hints can be substituted for ground truth hints during training) with some probability. Finally, the processed node embeddings $\mathbf{p}^t_{i}$ will be passed to the next step as the new recurrent state.

\subsection{Depth-First Search in CLRS-30}

Every algorithm is associated with a set of hints that should be predicted at each step. In CLRS-30, the DFS algorithm (Appendix \ref{app:dfsalg}) has the hints described in Table \ref{tab:dfshints}. In this algorithm, each hint is encoded for each node. For example, there is a pointer to the predecessor of every node in the graph. We call these types of hints \textit{per-node} hints as they have a value for every node. In other algorithms, a hint can be associated with the whole graph (a \textit{graph} hint). As an example, the \textit{min} pointer in binary search is a graph hint. It does not belong to any particular element of the input representation, but is instead shared between all of them---it is a property of the whole graph.

\begin{table}
    \begin{center}
      \caption{\textbf{Hints provided by CLRS-30 for the depth-first search algorithm.} All hints (except \textit{time}) are per-node hints. Hints with * correspond to a variable in the DFS algorithm given in Appendix \ref{app:dfsalg}.}
      \label{tab:dfshints}
      \begin{tabular}{clc}
        \toprule % <-- Toprule here
        \textbf{Hint} & \textbf{Explanation} & \textbf{Graph or Per-Node?}\\
        \midrule % <-- Midrule here
        $\pi_h$* & Pointer to predecessor for each node & Per-Node \\
        \textit{color}* & Color of each node & Per-Node \\
        $d$* & Time of discovery for each node & Per-Node \\
        $f$* & Time of finalization for each node & Per-Node \\
        $s_{\mathrm{prev}}$ & Pointer to previous node for each node & Per-Node \\
        $s$ & Current node & Per-Node \\
        $u$* & Node being explored & Per-Node \\
        $v$* & Node to be explored & Per-Node \\
        $s_{\mathrm{last}}$ & Last node explored for each node & Per-Node \\
        \textit{time}* & Time step & Graph\\
        \bottomrule
      \end{tabular}
    \end{center}
\end{table}

\section{Augmenting A GNN With A Stack}
\label{sec:gnn-with-call-stack}

\begin{figure*}
    \centering
    \includegraphics[width=0.8\linewidth]{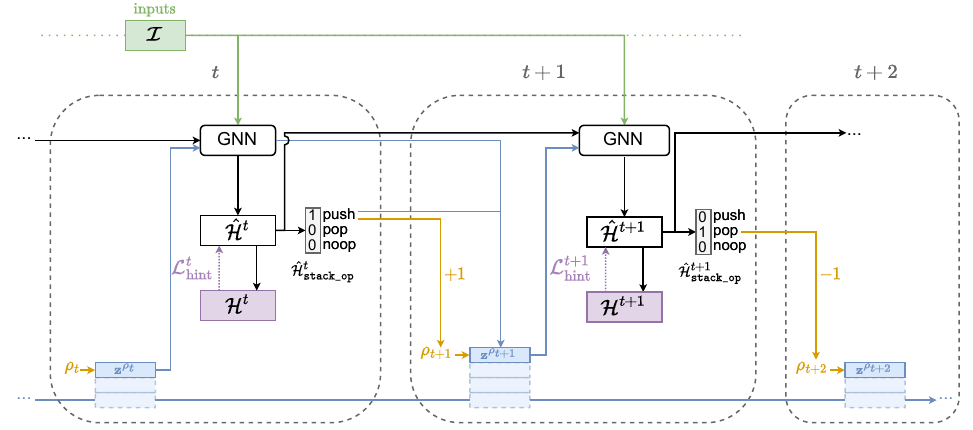}
    \caption{\textbf{Pushing and popping states with a stack-augmented GNN.} The input to the network is the state at the top of the stack, the hints, and the algorithm inputs. Note that we do not pass the recurrent state to the GNN (Section \ref{ssec:no-recurrent}). The network predicts the next processed features (which form the stack element), stack operation, and next hints. If the operation is a \push, the next stack element is formed and placed on the stack; if it is a \pop, the current top stack element is discarded.}
    \label{fig:method}
\end{figure*}

Recursive algorithms typically require storing state in a call stack, executing the recursive call, and finally restoring this state to complete the recursion step. To support similar reasoning, our method adds stack memory to the processor network described in Section \ref{ssec:CLRS}, enabling the network to push and pop latent states. In addition to arbitrary-sized read-write memory, this provides an inductive bias towards storing and recalling state like a call stack.

To allow the network to use the stack, we first add a one-hot encoded graph hint
$$
\hintstackop^t\in\{\push,\pop,\noop\footnote{Indicates no stack operation. The state of the stack is unmodified.}\}
$$
denoting the stack operation that the target algorithm performs at step $t$. The ground truth for this hint is $\push$ when entering a recursive call in the target algorithm, $\pop$ when returning from one, and $\noop$ otherwise. At inference time, when the network predicts this hint, latent states are pushed/popped from a stack, simulating a call stack during the execution of a recursive algorithm.

To utilize the stack operation predictions, we introduce a stack at step $t$ as $\mathcal{S}^t$, which is composed of a sequence of stack elements $\mathbf{z}^0, \dots,\mathbf{z}^{\pointer_t}$ where $\pointer_t$ indicates the number of elements on $\mathcal{S}^t$. We start with $\pointer_0:=0$ and define $\mathbf{z}^0 := \mathbf{0}$.

The elements that are pushed to the stack are defined by the type of stack. We introduce two types: a stack for every node whose elements are processed node features (a \textit{node-wise} stack), and a single stack for the graph for which elements are some pooled encoding of the node features (a \textit{graph} stack). Figure \ref{fig:method} provides a high-level demonstration of how the stack is used and how stack usage is learned. Stack operations are supervised such that the network learns to push and pop at precisely the same times as the recursive algorithm (i.e., when state needs to be saved, and when it needs to be recalled).

\subsection{Node-Wise Stack}
\label{ssec:nodelevel}
To store one element per node $i\in V$ we define $\mathbf{z}_i^{\pointer_t}\in\mathbb{R}^{d_\text{stack}}$ to be the top stack element corresponding to node $i$ at step $t$.
Depending on the predicted operation $\hat{\hints}_\texttt{stack\_op}^t$ we can then update the stack for step $t+1$ as follows:
\begin{align*}
    \pointer_{t+1} &=
    \begin{cases}
        \pointer_t + 1, \hspace{2pt} \mathrm{if} \hspace{2pt} \push\\
        \max\{\pointer_t-1,0\}), \hspace{2pt} \mathrm{if} \hspace{2pt} \pop\\
        \pointer_t, \hspace{2pt} \mathrm{if} \hspace{2pt} \noop
    \end{cases}\\
    \mathbf{z}_i^{\pointer_{t+1}} &=
    \begin{cases}
        \phi_\text{value}(\mathbf{p}_i^t), \hspace{2pt} \mathrm{if} \hspace{2pt} \push\\
        \mathbf{z}_i^{\change{\max\{}\pointer_{t-1}\change{,0\}}}, \hspace{2pt} \mathrm{if} \hspace{2pt} \pop\\
        \mathbf{z}_i^{\pointer_{t}}, \hspace{2pt} \mathrm{if} \hspace{2pt} \noop
    \end{cases}
\end{align*}
% \begin{itemize}
%     \item[] {\push:} $\pointer_{t+1}:=\pointer_t+1$ and $\mathbf{z}_i^{\pointer_{t+1}}:=\phi_\text{value}(\mathbf{p}_i^t)$
%     \item[] {\pop:} $\pointer_{t+1}:=\max\{\pointer_t-1,0\}$
%     \item[] {\noop:} $\pointer_{t+1}:=\pointer_t$
% \end{itemize}
Here, $\phi_\text{value}:\mathbb{R}^{d_\mathbf{h}}\rightarrow\mathbb{R}^{d_\text{stack}}$ denotes a (potentially learnable) function to decide which information to put on the stack. In each step $t$, we concatenate $\mathbf{z}^{\pointer_{t}}_i$ to the initial node embeddings that serve as input to the GNN $\psi$ given by $\mathbf{h}_i^{t}$ and optionally $\mathbf{p}_i^{t-1}$ (see Section \ref{ssec:no-recurrent}). 

Notably, using the top of the stack as an input to the network is effectively the same as providing a dynamic skip connection across time. As a result, we mitigate vanishing gradient issues because we do not need to backpropagate through intermediate time steps between when the state was first pushed and the current time.

%$\mathbf{h}_i^t:=f_n(\mathbf {x}_i^t)\>\|\>\mathbf{z}^{\pointer_t}_i$ is then concatenated to 

\subsection{Graph-Level Stack}
\label{ssec:pooling}
We also consider a graph-level stack. In this case, the stack element $\mathbf{z}^{\pointer_t}\in\mathbb{R}^{d_\text{stack}}$ is a vector of fixed size. The $\pop$ and $\noop$ operations are similar to the node-wise stack. In the case of a $\push$ operation, we update the stack with
\begin{equation}
    \mathbf{z}^{\pointer_{t+1}}:=\bigoplus_{i\in V}\phi_\text{value}(\mathbf{p}_i^t)
\end{equation}
where $\oplus$ is some permutation-invariant aggregation. In step $t$, the top stack element $\mathbf{z}^{\pointer_t}$ is  concatenated to the graph features $\mathbf{h}^t_g$. We use a 2-layer MLP for our value network $\phi_\text{value}$. As an alternative, we also explore taking the first $d_\text{stack}$ entries of the node embedding such that
\begin{equation}
    \label{eq:no-valuenet}
    \phi_\text{value}(\mathbf{p}_i^t):= (\mathbf{p}_i^t)_{0:d_\text{stack}}
\end{equation}
where $d_{h}\geq d_\text{stack}$.

\section{Stacks Are Not All You Need}
\label{sec:new-setup}

% \subsection{No State Recall Required With Per-Node Hints}
\subsection{Recursive Problems Require Additional Memory}
\label{ssec:graph-level-hints}

% Removed two sentences that I think are redundant and reduce readability:
% In many algorithms, there can never be more than one stack element corresponding to the same node. In the example of DFS, this is because each node can only be visited once. 
% As a result, the CLRS-30 algorithm is one with random access memory that has maximum size equal to the call stack since the maximum recursion depth in DFS grows linearly with the number of nodes in the graph.

\begin{figure}
    \centering
    \begin{subfigure}[b]{\linewidth}
         \centering
         \includegraphics[width=0.7\linewidth]{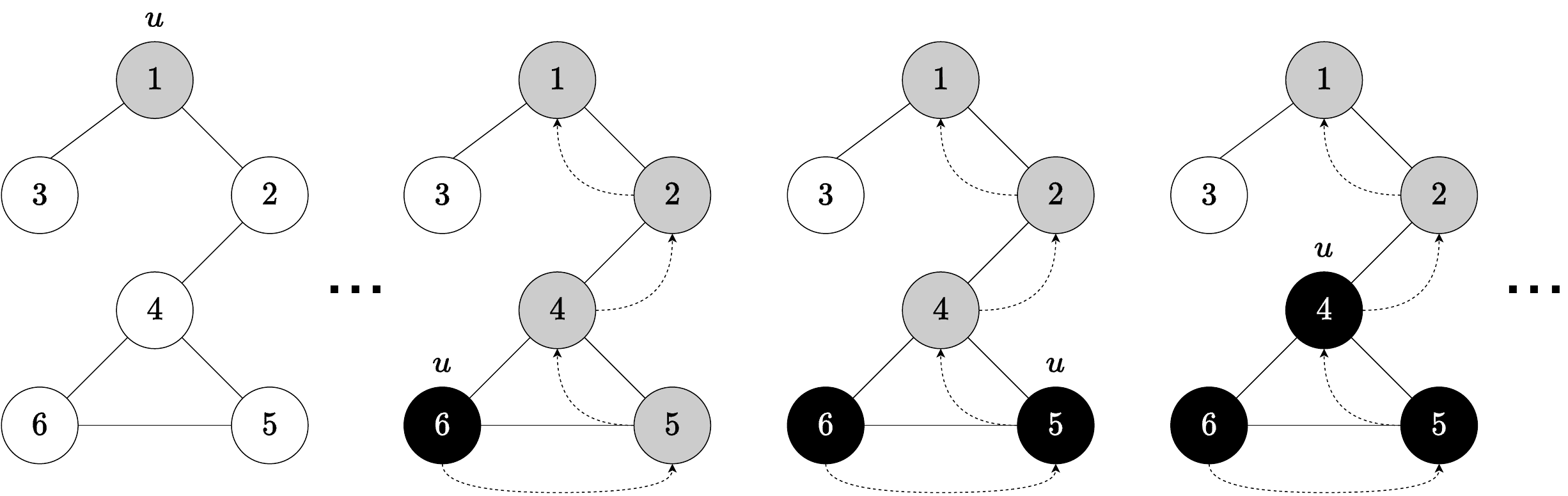}
         \caption{Per-node hints}
     \end{subfigure}
     \begin{subfigure}[b]{\linewidth}
         \centering
         \includegraphics[width=0.7\linewidth]{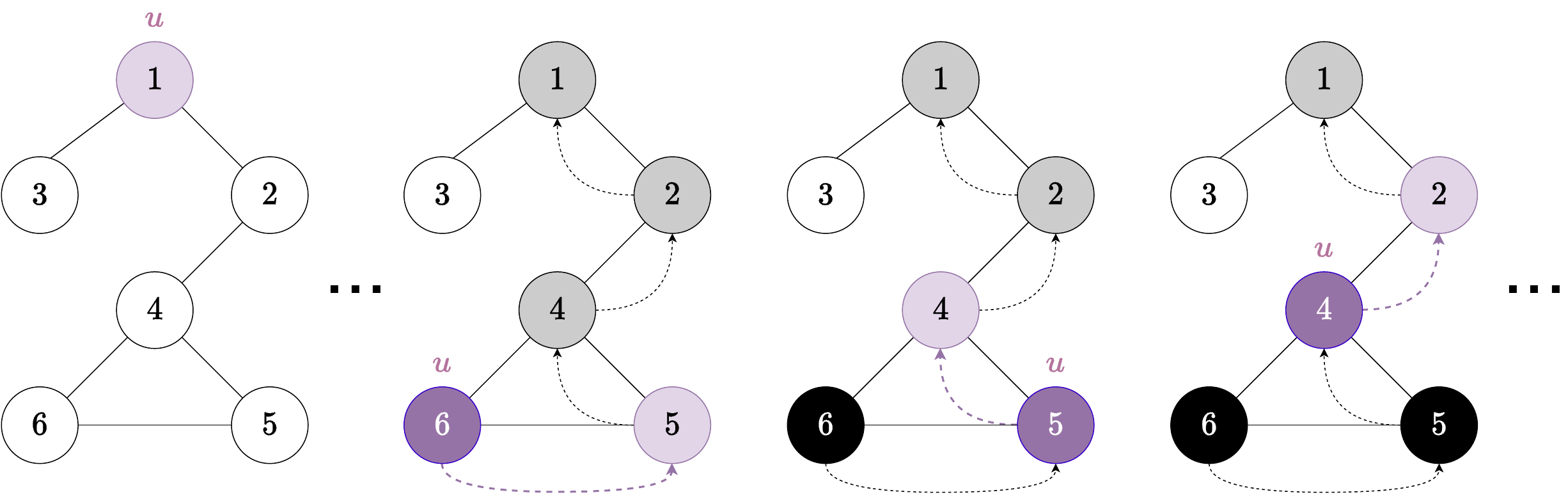}
         \caption{Graph hints (purple)}
     \end{subfigure}
    \caption{\textbf{Per-node hints are sufficient to determine the next step of DFS.} Some steps of the DFS algorithm are shown on a graph in (a). The nodes are colored white (unvisited), gray (being explored), or black (visited) according to the \textit{color} variable in DFS (see Appendix \ref{app:dfsalg}). The back-pointers indicate the predecessor of a node ($\pi$). The current focus of the algorithm is $u$. After reaching the state in the second step, the algorithm backtracks to the predecessor, Node 5. Finalizing Node 5 as it has no unexplored neighbors, the algorithm backtracks to Node 4 and so on. It knows which node to go to next, because it has the current node $u$, the predecessor pointers $\pi$ for all nodes, and the colors of all nodes. Therefore, per-node hints are enough to determine the next step of DFS without any additional state. In (b), we show how this would look with our proposed hints (Table \ref{tab:dfslocalhints}) which provide only information for $u$, the predecessor pointer for $u$, and the colors of all nodes. Only nodes and pointers colored in purple are known at each step. The predecessor pointer for $u$ is insufficient to backtrack more than one node as the algorithm can only backtrack from Node 6 to 5 before running out of purple-shaded back-pointers. The algorithm needs to recall the state relevant to Node 5 to backtrack further (as shown in the third step). Therefore, state recall is required to solve DFS with our hints.}
    \label{fig:dfssteps}
\end{figure}

The implementations of many recursive algorithms in CLRS-30 are not truly recursive. For instance, in DFS, hints, such as the predecessor $\pi$, are present for \textit{each} node (per-node hints). Together, these per-node hints provide \textit{global} information about the entire computation state. This information is already sufficient to deduce what to do in the next algorithm step \textit{without additional memory}. As demonstrated by Figure \ref{fig:dfssteps}, when per-node hints are used for critical variables in the algorithm, a stack is not required. Hence, the algorithm can be executed without recursive steps.

In contrast, the recursive implementation of DFS described by \citet{cormen_introduction_2009} (Appendix \ref{app:dfsalg}) uses variables that would be graph hints rather than per-node hints. For example, the algorithm has access to only the predecessor of the \textit{current} node rather than the predecessors of all nodes. The use of per-node hints in the CLRS-30 implementation is therefore problematic as it implies that the network will not be closely aligned with a recursive algorithm.

To remedy this, we propose using graph hints with the aim of achieving closer algorithmic alignment with recursion. As a demonstrative example, we modify the DFS implementation in CLRS-30 to use hints based directly on the variables in the algorithm (Table \ref{tab:dfslocalhints}), where all except \textit{color} are graph hints rather than per-node hints. These new hints are relative only to the current node being explored. The \textit{color} hint remains a per-node hint as this information is required when looping over the neighbors of a node in the algorithm (see line 5 of DFS-Visit() in Appendix \ref{app:dfsalg}). This configuration of hints is more similar to the state pushed to the call stack in a truly recursive DFS algorithm as we only save the state related to the node we are currently exploring. With this change, the network will not have enough information to execute DFS without storing and restoring state as the DFS algorithm does. A similar procedure can be applied to recursive algorithms in general by ensuring that variables which would be pushed onto the call stack in a recursive call are always graph hints.

\begin{table}[b]
    \begin{center}
      \caption{\textbf{Modified hints for the depth-first search algorithm.} All hints except \textit{color} are graph hints. All except \textit{stack\_op} correspond directly to the algorithm in Appendix \ref{app:dfsalg}.}
      \label{tab:dfslocalhints}
      \begin{tabular}{clc}
        \toprule % <-- Toprule here
        \textbf{Hint} & \textbf{Explanation} & \textbf{Graph or Per-Node?} \\
        \midrule % <-- Midrule here
        $u$ & Node being explored & Graph \\
        $u_\pi$ & Predecessor of node & Graph \\
        $u_d$ & Time when node discovered & Graph \\
        $u_f$ & Time when node finalized & Graph \\
        $u_v$ & Neighbor to be explored & Graph \\
        \textit{color} & Color of all nodes & Per-Node \\
        \textit{time} & Time step & Graph \\
        \textit{stack\_op} & Stack operation (\push/\pop/\noop) & Graph \\
        \bottomrule
      \end{tabular}
    \end{center}
\end{table}

\subsection{Recurrent States Can Encode Global Information}
% \subsection{Per-Node Hints Can Be Learned By A Recurrent State}
\label{ssec:no-recurrent}
Removing per-node hints is not enough as they can also be learned implicitly in the node-wise recurrent state $\mathbf{p}_i^{t-1}$. Whatever information can be carried from a hint, can also be learned as part of this hidden state. Therefore, the network could learn a representation similar to the previous unmodified hints through this state. In such a case, the network would not need to save or restore state once again. Consequently, we do not pass information in the recurrent state to the GNN and enforce that it relies only on the information on our stack. Hence, we modify Equation \ref{eq:processor} to
\begin{equation}
    \mathbf{p}_i^{t}, \mathbf{p}_{ij}^{t} = \psi(\mathbf{h}_i^t,  \mathbf{h}_{ij}^t, \mathbf{h}_g^t)%, \mathbf{z}^{\pointer_t})
\end{equation}
Similarly, a node-wise stack could act like a node-wise recurrent state if the stack were pushed to every step. In this case, the top of the stack would provide a recurrent state to the GNN which it could use to learn per-node hints implicitly. We avoid this problem as our stack is explicitly supervised through ground truth stack operations sampled from the target algorithm which would never require pushing state to the call stack in every step.

\subsection{Recursive Algorithms Generate Results Sequentially}
% \subsection{Results Are Generated Sequentially in Recursive Algorithms}
\label{ssec:collection}

\begin{figure*}[b]
    \centering
    \includegraphics[width=0.7\textwidth]{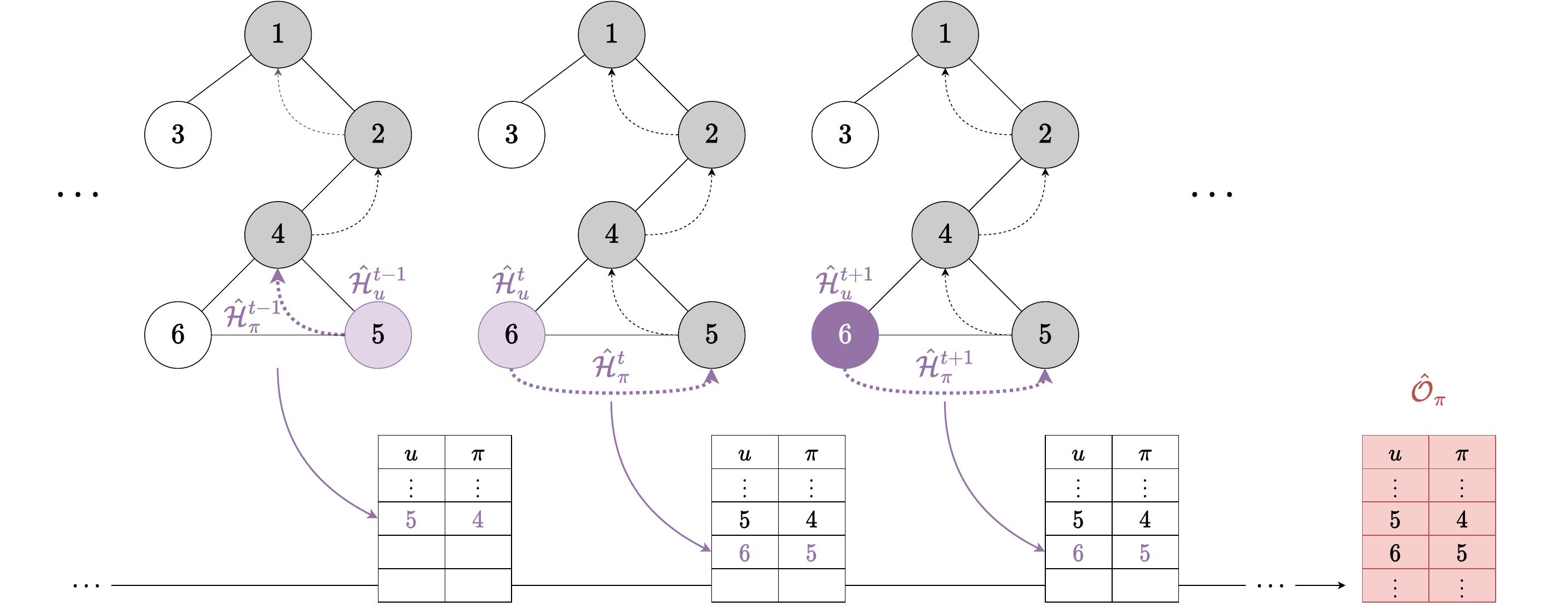}
    \caption{\textbf{Collecting the output from predecessor hint predictions.} In each step, the network only predicts graph hints like the current node $\hat{\mathcal{H}}_u^t$ and its predecessor $\hat{\mathcal{H}}_\pi^t$ marked in purple (dark for black nodes, light for gray nodes). To collect the final output, we maintain a table during execution. In each step, we take the entry of the predicted current node $\hat{\mathcal{H}}_u^t$ and overwrite its predecessor with the predicted predecessor $\hat{\mathcal{H}}_\pi^t$. Predecessors of nodes that are not purple are only shown for context. Our network does not have access to the corresponding hints.}
    \label{fig:collect}
\end{figure*}

In CLRS-30, the complete output of the algorithm (e.g., the sequence of nodes in a search) is predicted at once from the processed features. This is not how recursive algorithms typically work. A DFS algorithm, for example, will generate the result by outputting each new node found as the search explores the graph. Moreover, since we use the top stack element and node embeddings to predict the output of the algorithm, to predict the output in the way that CLRS-30 does it, we need to memorize all the nodes found by the search as part of the stack element, introducing a memory bottleneck that will degrade performance as the number of nodes in the graph grows. Therefore, we modify CLRS-30 to tabulate predicted outputs for each node as the network executes the algorithm and, at the end, use this table to collect these results together into a single final output. This is more like how a typical recursive algorithm generates its result. We provide an example of this for DFS in Figure \ref{fig:collect}. This method can also be applied to other recursive algorithms. For example, in quicksort, the final position of the pivot element can be tabulated after each recursive call.

\section{Results \& Discussion}
\label{sec:results}

\begin{table*}
\centering
 \caption{\textbf{Call stacks improve generalization performance.} Test accuracy is measured for graphs with 32 (in-distribution) and 96 (out-of-distribution) nodes respectively. The train and validation set contain a mix of graphs of up to 32 nodes as described in Appendix \ref{sec:dataset}. \textit{Ours} denotes the setting where hints are mostly graph hints (Section \ref{ssec:graph-level-hints}), with various ablations. We report the mean and standard deviation of test accuracy over three runs at the point of best validation accuracy (early stopping). For complete implementation details, see Appendix \ref{app:impdetails}. Our node-wise call stack configuration achieves the best out-of-distribution generalization performance.} % your caption here
\label{tab:results}
\begin{tabular}{l>{(\refstepcounter{rowno}\therowno)}cccccccccc}
\multicolumn{2}{@{}c}{} & \rot{Graph-Level stack (Sec. \ref{ssec:pooling})} & \rot{Node-Wise stack (Sec. \ref{ssec:nodelevel})}  & \rot{Hidden state $\mathbf{p}_i^{t-1}$ (Sec. \ref{ssec:no-recurrent})} & \rot{Output collection (Sec. \ref{ssec:collection})} & \rot{Teacher forcing (50\%)} & \rot{$\phi_\text{value}$ learned (Eq. \ref{eq:no-valuenet})} & \rot{Attention (App. \ref{app:attention})} & \rot{Test Acc. (32 nodes)} & \rotatebox{45}{Test Acc. (96 nodes)} \\
\toprule
\multirow{1}{*}{\citeauthor{ibarz_generalist_2022}} & \label{row:dfs_orig} & N/A & N/A & \graycheck & \graycross & \graycheck & N/A & N/A & \confint{99.79}{0.20} & \confint{53.92}{14.06} \\
\hline
\multirow{11}{*}{Ours} & \label{row:ours} & \graycheck & \graycross & \graycross & \graycheck & \graycheck & \graycheck & \graycross & \confint{98.00}{0.60} & \confint{73.00}{6.31}\\
                      & \label{row:no-stack} & \blackcross & \graycross & \graycross & \graycheck & \graycheck & N/A & N/A & \confint{65.33}{4.69} & \confint{72.88}{7.47}\\
                      & \label{row:recurrent} & \graycheck & \graycross & \blackcheck & \graycheck & \graycheck & \graycheck & \graycross & \confint[b]{100.00}{0.00} & \confint{82.19}{1.07} \\
                      & \label{row:recurrent-no-stack} & \blackcross & \graycross & \blackcheck & \graycheck & \graycheck & N/A & N/A & \confint[b]{100.00}{0.00} & \confint{78.65}{5.44}\\
                      & \label{row:no-collection} & \graycheck & \graycross & \graycross & \blackcross & \graycheck & \graycheck & \graycross & \confint{50.54}{3.90} & \confint{25.08}{0.97}\\
                      & \label{row:no-tf} & \graycheck & \graycross & \graycross & \graycheck & \blackcross & \graycheck & \graycross & \confint{71.73}{2.14} & \confint{43.88}{15.73}\\
                      & \label{row:no-valuenet} & \graycheck & \graycross & \graycross & \graycheck & \graycheck & \blackcross & \graycross & \confint{97.54}{0.83} & \confint{67.42}{3.67}\\
                      & \label{row:attention} & \graycheck & \graycross & \graycross & \graycheck & \graycheck & \graycheck & \blackcheck & \confint{92.27}{5.04} & \confint{49.40}{2.28}\\
                      & \label{row:nodelevel} & \blackcross & \blackcheck & \graycross & \graycheck & \graycheck & \graycheck & N/A & \confint[b]{100.00}{0.00} & \confint[b]{100.00}{0.00}\\
                      & \label{row:nodelevel-recurrent} & \blackcross & \blackcheck & \blackcheck & \graycheck & \graycheck & \graycheck & N/A & \confint[b]{100.00}{0.00} & \confint{99.79}{0.29}\\
\bottomrule
\end{tabular}
\end{table*}

In Table \ref{tab:results}, we summarize our results on different network configurations. We refer to test accuracy on larger graphs (96 nodes) as out-of-distribution (OOD) performance. While the setup proposed by \citet{ibarz_generalist_2022} (Experiment \ref{row:dfs_orig}) achieves near-perfect test accuracy on our dataset, it fails to generalize to larger graphs. In contrast, our method achieves similar in-distribution performance, and at 73\% accuracy for a graph-level stack and 100\% for a node-wise stack, \textit{drastically better OOD generalization performance} (Experiment \ref{row:ours} and \ref{row:nodelevel}) while using \textit{less memory} (Appendix \ref{app:runtime}). 

Experiment \ref{row:no-tf} shows that teacher forcing is required to achieve the generalization performance we see. In line with a similar observation by \citet{ibarz_generalist_2022}, DFS is one of the algorithms in CLRS-30 that benefits from teacher forcing.

% We believe this is because teacher forcing helps alleviate the effect of errors in the initial iterations of training when learning the stack operation hint.

In Experiment \ref{row:no-stack}, we note that removing the stack only has a minor effect on the the OOD performance but significantly reduces in-distribution accuracy. We hypothesize that this is the performance that can be achieved when only the hints are provided. There is no additional information propagation in terms of stack or hidden state. The main generalization benefit stems from turning our per-node hints into graph hints and collecting outputs as described in Section \ref{sec:new-setup}. This is a result of a significant improvement in alignment with the original DFS algorithm.

Notably, not learning the value network $\phi_\mathrm{value}$ (Equation \ref{eq:no-valuenet}), does not noticeably impact performance (Experiment \ref{row:no-valuenet}). This could indicate that learning an encoding of the processed features is unnecessary as the GNN $\psi$, which processes features, is able to learn an effective encoding by itself. In addition, as demonstrated by Experiment \ref{row:no-collection}, collecting the outputs is a crucial component of aligning with the DFS algorithm as it has a significant impact on accuracy.

While we adjusted the DFS hints and output collection in a way that theoretically makes a graph-level stack sufficient, learning to store the relevant information for each node appears to be easier than also learning which node to focus on. This reflects in the fact that adding a (per-node) recurrent state to the graph-level stack (similar to \citet{ibarz_generalist_2022} in Experiment \ref{row:dfs_orig}) yields another significant boost in performance (Experiment \ref{row:recurrent}). Removing the graph-level stack only has a minor effect (Experiment \ref{row:recurrent-no-stack}). %\todo{Mention the fact that there is no need to learn which node we need to focus on at each step with a node-wise stack. No aggregation so no need to learn to disentangle. No need to learn which one leave out either. Easier to learn.}

Based on this insight, we evaluate a network augmented with a node-wise stack as described in Section \ref{ssec:nodelevel}. We use the same graph hints and outputs as described in Section \ref{sec:new-setup}. This allows us to propagate per-node information while maintaining the inductive bias of a call stack. \textbf{We achieve perfect test accuracy in-distribution as well as on larger out-of-distribution graphs} with this modification (Experiment \ref{row:nodelevel}). When reintroducing the recurrent state, the network learns to use the node-wise call stack and achieves only slightly worse generalization accuracy (Experiment \ref{row:nodelevel-recurrent}).

As our network is essentially a graph RNN, it faces the same forgetting issues as other recurrent networks \cite{ClockworkRNN,hochreiter_untersuchungen_1991}. The stack resolves these forgetting issues, which are caused by using a node-wise hidden state instead of a stack as in Experiment \ref{row:recurrent-no-stack}. Our method enables the network to perfectly recall state from more than one step in the past---a particularly practical feature when dealing with problems of high recursion depth.

% The fact that we outperform a recurrent state when we use a stack hints to the fact that we avoid the memory bottleneck of the hidden state in recurrent networks. This is because we can 

Our modifications make the implementation in CLRS-30 significantly more closely aligned with the recursive DFS algorithm defined by \citet{cormen_introduction_2009}. How has previous work also achieved strong empirical \textit{in-distribution} results on DFS without this close alignment? Noting that memoization does not change expressiveness, DFS can be expressed as a dynamic programming (DP) problem. Given the structural alignment between GNNs and DP (shown by \citet{xu_what_2020}), in prior work, the hints were sufficient to allow the algorithm to be solved as DP. It is perhaps because other approaches do not conform as closely to DP as our approach structurally aligns with recursion that these implementations do not result in networks which generalize as well as ours \textit{out-of-distribution}.

%\todo{Maybe we should make \textit{Ours} the one \textbf{WITH} hidden state (as it always seems to perform better) however, then our experimental design would look a little weird as we do all ablations (no value net, attention, no teacher forcing, \dots) without. So I guess it would be best to just describe why it performs better with hidden state}

% \todo{
% \begin{itemize}
%     \item 
%     \begin{itemize}
%         \item main benefit here seems to be that our setup is better aligned with actual DFS (everything local instead of global) as run \ref{row:no-stack} only performs marginally worse in terms of generalization (although significantly better in distribution, WHY?)
%         \item when comparing with DFS baseline from generalist, note that we don't have a per-node hidden state as they have in DFS which makes us more aligned but harder to learn
%     \end{itemize}
%     \item No collection has to fail because we need to fit information about variable number of nodes in fixed-size stack embedding (which is the only hidden state we can transfer from time step to time step)
%     \item 
%     \item Teacher forcing is necessary (run \ref{row:no-tf})
%     \item \ref{row:attention} and \ref{row:no-valuenet} don't need to be mentioned in my opinion, they already are in the respective sections.
%     \item Node-level: in comparison to run \ref{row:recurrent-no-stack} we have
% \end{itemize}
% }

\section{Related Work}
\label{sec:related}
Motivated by patterns which are difficult to learn in deep neural networks, \citet{joulin_inferring_2015} developed stack-augmented recurrent networks. They use the stack to learn control over the memory of the network, enabling it to learn with infinite structured memory. They showed that these networks are able to learn some basic algorithms (such as binary addition) which require memorization. However, they did not study recursive algorithms or attempt to align the learned method with the algorithm structure. \citet{cai_making_2017} incorporated recursion through the Neural Programmer-Interpreter (NPI) framework \cite{reed_neural_2016}. This was achieved by incorporating recursive elements into the NPI traces (somewhat similar to incorporating per-node hints in CLRS-30). They demonstrated strong generalization performance when learning sorting algorithms. In contrast to our approach, they do not explicitly introduce a call stack to learn the relevant state. \citet{petersen_learning_2021} proposed a method of relaxing conditions on control structures in algorithms such that they were smoothly differentiable. This approach allows neural networks to directly learn the relaxed algorithms. Similar to the method proposed by \citet{cai_making_2017}, this technique also cannot permit reasoning like recursive algorithms as there is no saving or restoring of state.

\section{Conclusion \& Future Work}
\label{sec:conclusion}

To enable neural networks to inherit some of the generalization properties of recursive algorithms, we introduced a new framework for augmenting GNNs with a call stack. This framework permits GNNs to execute recursive algorithms. We also proposed changes for sampling intermediate algorithm trajectories and predicting outputs in the CLRS-30 algorithmic reasoning benchmark that improved structural alignment with recursion. With these improvements, our framework allowed a GNN with a call stack to significantly outperform previous work when generalizing out-of-distribution on DFS. Moreover, our stack-augmented graph neural network has the ability to perfectly recall state from history, avoiding the memory bottleneck of hidden states in recurrent networks.

% Some of our attempted modifications, such as the attention mechanism, did not improve our results. We believe a more detailed analysis of these settings could allow gains to be made. With attention, mean-pooling could enable the neural network to generalize more easily out-of-distribution as the magnitude of the encoded graph features will be independent of graph size. 
% A useful insight may arise from analyzing the accuracy of stack predictions and visualizing our hints. This could lead to a better understanding of what our algorithmically-aligned approach does. It would allow us to verify that the network is acting like a recursive algorithm. 
% In addition, to confirm the generality of our results, further experiments on other recursive algorithms (e.g., quicksort) are required. 
One goal of future work would be to formalize the structural alignment of our modifications in CLRS-30 with our stack-augmented GNN architecture, similar to the alignment between DP and GNNs shown by \citet{xu_what_2020}. This could pave the way for further insights about reasoning recursively with neural networks. We also note that our method relies on ground-truth hints to supervise stack usage. This is usually not available outside an algorithmic reasoning setting. Usage of the stack could instead be learned through reinforcement learning, supervising the stack with the policy loss. This would enable call stacks to be used with neural networks beyond only known recursive algorithms. Similar to the work by \citet{li_strong_2020}, it could allow the network to discover new methods of using the stack which can outperform known solutions. For example, the policy network of a navigation agent could employ our architecture. This could enable the agent to better learn to map and navigate its environment as this task requires recursively planning paths.

Nevertheless, in its current state, our work is the first to demonstrate the use of a stack-augmented neural network in NAR. We have used this architecture to improve algorithmic alignment in recursive problems and enlarged the class of algorithms we can precisely reason about with GNNs. As a result, this work is a step towards transferable algorithmic knowledge \cite{xhonneux_how_2021} and generalist algorithmic learners \cite{ibarz_generalist_2022}. We hope that our work sheds light on future directions in the journey towards learning to reason with neural networks.

% \section*{Author Contributions}
% Authors of accepted papers are \emph{encouraged} to include a statement that declares the individual contribution of every author, especially when there are co-authors that made equal contributions to the research.
% You may adopt the \href{https://credit.niso.org/}{Contributor Roles Taxonomy (CRediT)} methodology for attributing contributions.
% Do not include this section in the version for blind review.
% This section does not count towards the page limit.

% ONLY in camera-ready version
\section*{Acknowledgements}
DJ and JJ sincerely thank Edan Toledo for valuable technical contributions which did not appear in the final manuscript, Dobrik Georgiev for assistance with CLRS-30, and Yonatan Gideoni for reviewing early drafts of this paper. All the authors thank Zhe Wang and Murray Shanahan for reviewing the final draft, and finally, all reviewers for their detailed and insightful comments which helped significantly improve this work.

% \textcolor{red}{Finally, we would like to thank Reviewer \#2 at the ICML 2023 Workshop on Knowledge and Logical Reasoning in the Era of Data-Driven Learning for their detailed and insightful comments on our work.}

% For natbib users:
\bibliographystyle{unsrtnat}
\bibliography{main}

\begin{thebibliography}{19}
\providecommand{\natexlab}[1]{#1}
\providecommand{\url}[1]{\texttt{#1}}
\expandafter\ifx\csname urlstyle\endcsname\relax
  \providecommand{\doi}[1]{doi: #1}\else
  \providecommand{\doi}{doi: \begingroup \urlstyle{rm}\Url}\fi

\bibitem[Xu et~al.(2020)Xu, Li, Zhang, Du, Kawarabayashi, and
  Jegelka]{xu_what_2020}
Keyulu Xu, Jingling Li, Mozhi Zhang, Simon~S. Du, Ken{-}ichi Kawarabayashi, and
  Stefanie Jegelka.
\newblock What can neural networks reason about?
\newblock In \emph{8th International Conference on Learning Representations,
  {ICLR} 2020, Addis Ababa, Ethiopia, April 26-30, 2020}. OpenReview.net, 2020.
\newblock URL \url{https://openreview.net/forum?id=rJxbJeHFPS}.

\bibitem[Veličković and Blundell(2021)]{velickovic_neural_2021}
Petar Veličković and Charles Blundell.
\newblock Neural algorithmic reasoning.
\newblock \emph{Patterns}, 2\penalty0 (7):\penalty0 100273, July 2021.
\newblock ISSN 2666-3899.
\newblock \doi{10.1016/j.patter.2021.100273}.
\newblock URL
  \url{https://www.sciencedirect.com/science/article/pii/S2666389921000994}.

\bibitem[Li et~al.(2020)Li, Gimeno, Kohli, and Vinyals]{li_strong_2020}
Yujia Li, Felix Gimeno, Pushmeet Kohli, and Oriol Vinyals.
\newblock Strong {Generalization} and {Efficiency} in {Neural} {Programs}, July
  2020.
\newblock URL \url{http://arxiv.org/abs/2007.03629}.
\newblock arXiv:2007.03629 [cs, stat].

\bibitem[Numeroso et~al.(2023)Numeroso, Bacciu, and
  Veli{\v{c}}kovi{\'c}]{numeroso2023dual}
Danilo Numeroso, Davide Bacciu, and Petar Veli{\v{c}}kovi{\'c}.
\newblock Dual algorithmic reasoning.
\newblock In \emph{The Eleventh International Conference on Learning
  Representations}, 2023.
\newblock URL \url{https://openreview.net/forum?id=hhvkdRdWt1F}.

\bibitem[Velickovic et~al.(2022)Velickovic, Badia, Budden, Pascanu, Banino,
  Dashevskiy, Hadsell, and Blundell]{velickovic_clrs_2022}
Petar Velickovic, Adri{\`{a}}~Puigdom{\`{e}}nech Badia, David Budden, Razvan
  Pascanu, Andrea Banino, Misha Dashevskiy, Raia Hadsell, and Charles Blundell.
\newblock The {CLRS} algorithmic reasoning benchmark.
\newblock In Kamalika Chaudhuri, Stefanie Jegelka, Le~Song, Csaba
  Szepesv{\'{a}}ri, Gang Niu, and Sivan Sabato, editors, \emph{International
  Conference on Machine Learning, {ICML} 2022, 17-23 July 2022, Baltimore,
  Maryland, {USA}}, volume 162 of \emph{Proceedings of Machine Learning
  Research}, pages 22084--22102. {PMLR}, 2022.
\newblock URL \url{https://proceedings.mlr.press/v162/velickovic22a.html}.

\bibitem[Del{\'{e}}tang et~al.(2023)Del{\'{e}}tang, Ruoss, Grau{-}Moya,
  Genewein, Wenliang, Catt, Cundy, Hutter, Legg, Veness, and
  Ortega]{DeltangChomskyHierarchy}
Gr{\'{e}}goire Del{\'{e}}tang, Anian Ruoss, Jordi Grau{-}Moya, Tim Genewein,
  Li~Kevin Wenliang, Elliot Catt, Chris Cundy, Marcus Hutter, Shane Legg, Joel
  Veness, and Pedro~A. Ortega.
\newblock Neural networks and the chomsky hierarchy.
\newblock In \emph{The Eleventh International Conference on Learning
  Representations, {ICLR} 2023, Kigali, Rwanda, May 1-5, 2023}. OpenReview.net,
  2023.
\newblock URL \url{https://openreview.net/pdf?id=WbxHAzkeQcn}.

\bibitem[Cormen et~al.(2009)Cormen, Leiserson, Rivest, and
  Stein]{cormen_introduction_2009}
Thomas~H. Cormen, Charles~E. Leiserson, Ronald~L. Rivest, and Clifford Stein.
\newblock \emph{Introduction to algorithms}.
\newblock MIT Press, Cambridge, Massachusetts, 3rd edition, 2009.
\newblock ISBN 978-0-262-03384-8 978-0-262-53305-8.

\bibitem[Bevilacqua et~al.(2023)Bevilacqua, Nikiforou, Ibarz, Bica, Paganini,
  Blundell, Mitrovic, and Velickovic]{bevilacqua_neural_2023}
Beatrice Bevilacqua, Kyriacos Nikiforou, Borja Ibarz, Ioana Bica, Michela
  Paganini, Charles Blundell, Jovana Mitrovic, and Petar Velickovic.
\newblock Neural algorithmic reasoning with causal regularisation.
\newblock In Andreas Krause, Emma Brunskill, Kyunghyun Cho, Barbara Engelhardt,
  Sivan Sabato, and Jonathan Scarlett, editors, \emph{International Conference
  on Machine Learning, {ICML} 2023, 23-29 July 2023, Honolulu, Hawaii, {USA}},
  volume 202 of \emph{Proceedings of Machine Learning Research}, pages
  2272--2288. {PMLR}, 2023.
\newblock URL \url{https://proceedings.mlr.press/v202/bevilacqua23a.html}.

\bibitem[Ibarz et~al.(2022)Ibarz, Kurin, Papamakarios, Nikiforou, Bennani,
  Csord{\'{a}}s, Dudzik, Bosnjak, Vitvitskyi, Rubanova, Deac, Bevilacqua,
  Ganin, Blundell, and Velickovic]{ibarz_generalist_2022}
Borja Ibarz, Vitaly Kurin, George Papamakarios, Kyriacos Nikiforou, Mehdi
  Bennani, R{\'{o}}bert Csord{\'{a}}s, Andrew~Joseph Dudzik, Matko Bosnjak,
  Alex Vitvitskyi, Yulia Rubanova, Andreea Deac, Beatrice Bevilacqua, Yaroslav
  Ganin, Charles Blundell, and Petar Velickovic.
\newblock A generalist neural algorithmic learner.
\newblock In Bastian Rieck and Razvan Pascanu, editors, \emph{Learning on
  Graphs Conference, LoG 2022, 9-12 December 2022, Virtual Event}, volume 198
  of \emph{Proceedings of Machine Learning Research}, page~2. {PMLR}, 2022.
\newblock URL \url{https://proceedings.mlr.press/v198/ibarz22a.html}.

\bibitem[Koutn{\'{\i}}k et~al.(2014)Koutn{\'{\i}}k, Greff, Gomez, and
  Schmidhuber]{ClockworkRNN}
Jan Koutn{\'{\i}}k, Klaus Greff, Faustino~J. Gomez, and J{\"{u}}rgen
  Schmidhuber.
\newblock A clockwork {RNN}.
\newblock In \emph{Proceedings of the 31th International Conference on Machine
  Learning, {ICML} 2014, Beijing, China, 21-26 June 2014}, volume~32 of
  \emph{{JMLR} Workshop and Conference Proceedings}, pages 1863--1871.
  JMLR.org, 2014.
\newblock URL \url{http://proceedings.mlr.press/v32/koutnik14.html}.

\bibitem[Hochreiter(1991)]{hochreiter_untersuchungen_1991}
Sepp Hochreiter.
\newblock Untersuchungen zu dynamischen neuronalen {Netzen}.
\newblock \emph{Master's thesis, Institut fur Informatik, Technische
  Universitat, Munchen}, 1:\penalty0 1--150, 1991.

\bibitem[Joulin and Mikolov(2015)]{joulin_inferring_2015}
Armand Joulin and Tomas Mikolov.
\newblock Inferring {Algorithmic} {Patterns} with {Stack}-{Augmented}
  {Recurrent} {Nets}.
\newblock In \emph{Advances in {Neural} {Information} {Processing} {Systems}},
  volume~28. Curran Associates, Inc., 2015.

\bibitem[Cai et~al.(2017)Cai, Shin, and Song]{cai_making_2017}
Jonathon Cai, Richard Shin, and Dawn Song.
\newblock Making neural programming architectures generalize via recursion.
\newblock In \emph{5th International Conference on Learning Representations,
  {ICLR} 2017, Toulon, France, April 24-26, 2017, Conference Track
  Proceedings}. OpenReview.net, 2017.
\newblock URL \url{https://openreview.net/forum?id=BkbY4psgg}.

\bibitem[Reed and de~Freitas(2016)]{reed_neural_2016}
Scott~E. Reed and Nando de~Freitas.
\newblock Neural programmer-interpreters.
\newblock In Yoshua Bengio and Yann LeCun, editors, \emph{4th International
  Conference on Learning Representations, {ICLR} 2016, San Juan, Puerto Rico,
  May 2-4, 2016, Conference Track Proceedings}, 2016.
\newblock URL \url{http://arxiv.org/abs/1511.06279}.

\bibitem[Petersen et~al.(2021)Petersen, Borgelt, Kuehne, and
  Deussen]{petersen_learning_2021}
Felix Petersen, Christian Borgelt, Hilde Kuehne, and Oliver Deussen.
\newblock Learning with algorithmic supervision via continuous relaxations.
\newblock In Marc'Aurelio Ranzato, Alina Beygelzimer, Yann~N. Dauphin, Percy
  Liang, and Jennifer~Wortman Vaughan, editors, \emph{Advances in Neural
  Information Processing Systems 34: Annual Conference on Neural Information
  Processing Systems 2021, NeurIPS 2021, December 6-14, 2021, virtual}, pages
  16520--16531, 2021.
\newblock URL
  \url{https://proceedings.neurips.cc/paper/2021/hash/89ae0fe22c47d374bc9350ef99e01685-Abstract.html}.

\bibitem[Xhonneux et~al.(2021)Xhonneux, Deac, Veličković, and
  Tang]{xhonneux_how_2021}
Louis-Pascal Xhonneux, Andreea-Ioana Deac, Petar Veličković, and Jian Tang.
\newblock How to transfer algorithmic reasoning knowledge to learn new
  algorithms?
\newblock In \emph{Advances in {Neural} {Information} {Processing} {Systems}},
  volume~34, pages 19500--19512. Curran Associates, Inc., 2021.

\bibitem[Durrett(2006)]{durrett_2006}
Rick Durrett.
\newblock \emph{Random Graph Dynamics}.
\newblock Cambridge Series in Statistical and Probabilistic Mathematics.
  Cambridge University Press, 2006.
\newblock \doi{10.1017/CBO9780511546594}.

\bibitem[Gilmer et~al.(2017)Gilmer, Schoenholz, Riley, Vinyals, and
  Dahl]{gilmer2017neural}
Justin Gilmer, Samuel~S Schoenholz, Patrick~F Riley, Oriol Vinyals, and
  George~E Dahl.
\newblock Neural message passing for quantum chemistry.
\newblock In \emph{International conference on machine learning}, pages
  1263--1272. PMLR, 2017.

\bibitem[Kingma and Ba(2015)]{DBLP:journals/corr/KingmaB14}
Diederik~P. Kingma and Jimmy Ba.
\newblock Adam: {A} method for stochastic optimization.
\newblock In Yoshua Bengio and Yann LeCun, editors, \emph{3rd International
  Conference on Learning Representations, {ICLR} 2015, San Diego, CA, USA, May
  7-9, 2015, Conference Track Proceedings}, 2015.
\newblock URL \url{http://arxiv.org/abs/1412.6980}.

\end{thebibliography}
% For bibLaTeX users:
% \printbibliography

\appendix

\newpage

\section{\change{Altering The} Dataset With \change{Balanced Recursion Branches}}
\label{sec:dataset}
The graph distribution in CLRS-30 for training and testing is generated from randomly sampling Erdős–Rényi (E-R) graphs with different edge connection probabilities. The expected distance between two nodes in these graphs is logarithmic in the number of nodes \cite{durrett_2006}. \change{To also study balanced recursion branches}, we modify the graph distribution by replacing 15\% with randomly generated binary trees. \change{These graphs have branching chains that are in expectation the same length, making them different to E-R graphs}. % As a result, the network will be required to recall states \change{from one branch before and then explore the next branch, saving a similar number of states.}

\begin{figure}[h]
     \centering
     \hfill
     \begin{subfigure}[b]{0.22\textwidth}
         \centering
         \includegraphics[width=0.9\textwidth]{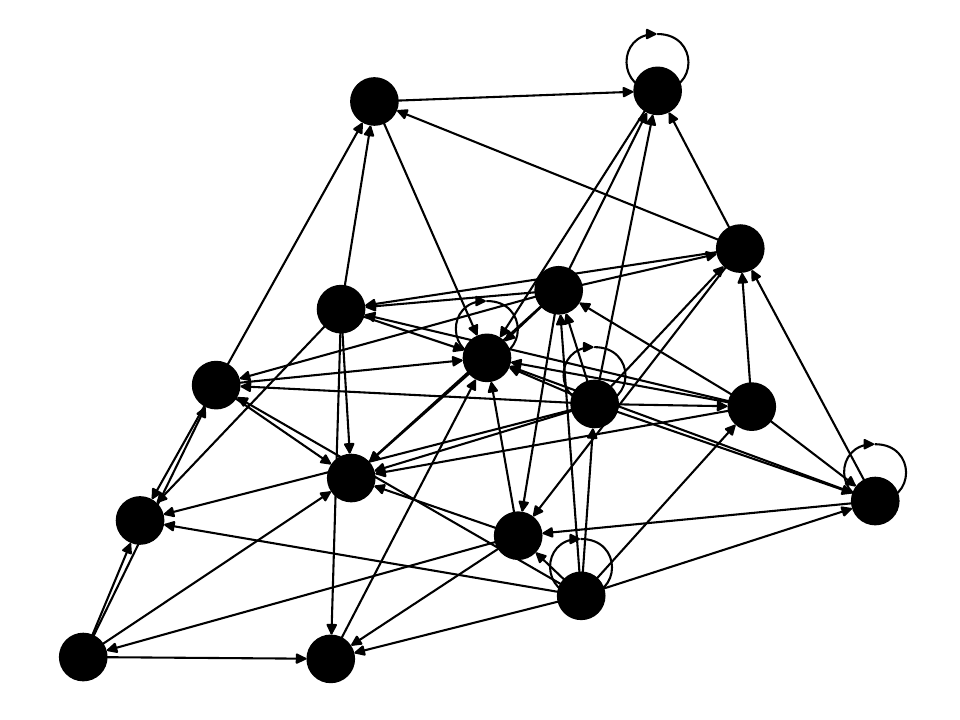}
         \caption{E-R, $p=0.3$}
     \end{subfigure}
     \hfill
     \begin{subfigure}[b]{0.22\textwidth}
         \centering
         \includegraphics[width=0.9\textwidth]{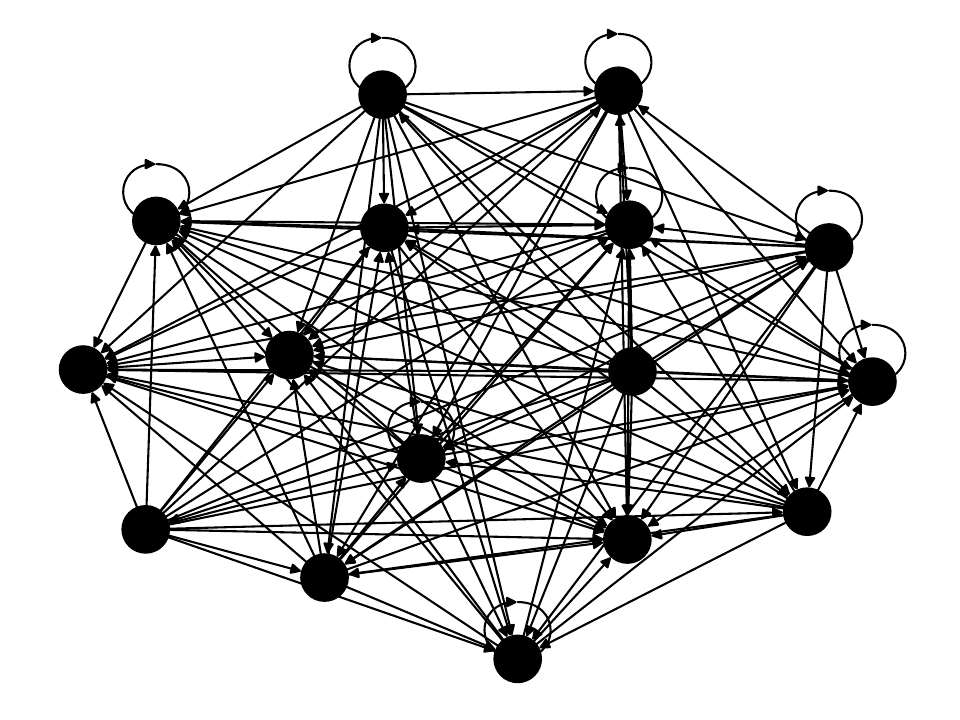}
         \caption{E-R, $p=0.8$}
     \end{subfigure}
     \hfill
     \begin{subfigure}[b]{0.22\textwidth}
         \centering
         \includegraphics[width=0.9\textwidth]{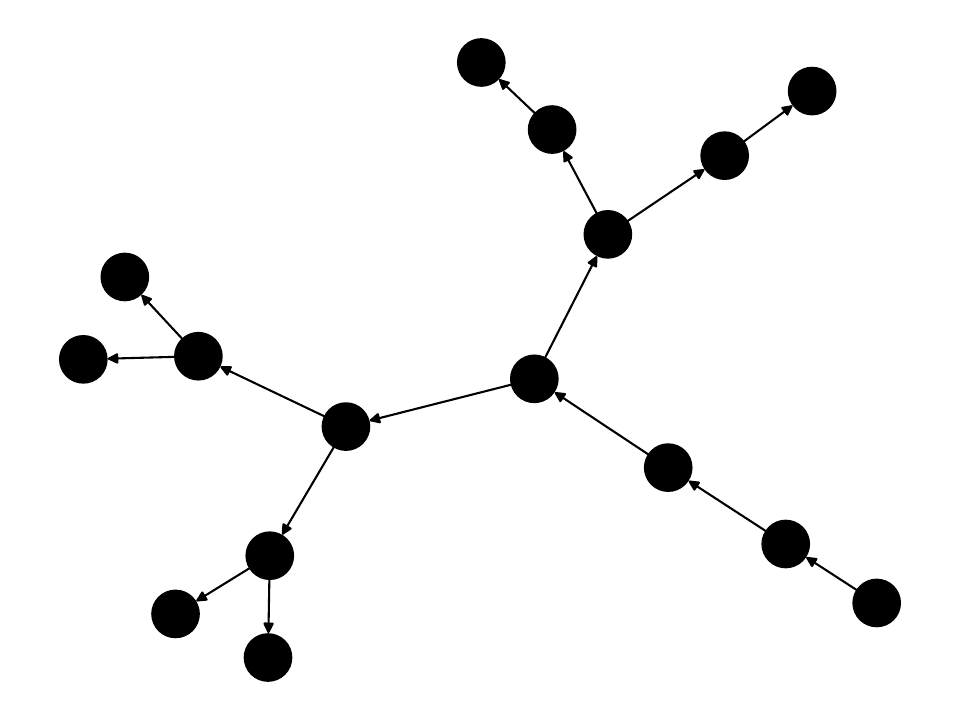}
         \caption{Binary Tree 1}
     \end{subfigure}
     \hfill
     \begin{subfigure}[b]{0.22\textwidth}
         \centering
         \includegraphics[width=0.9\textwidth]{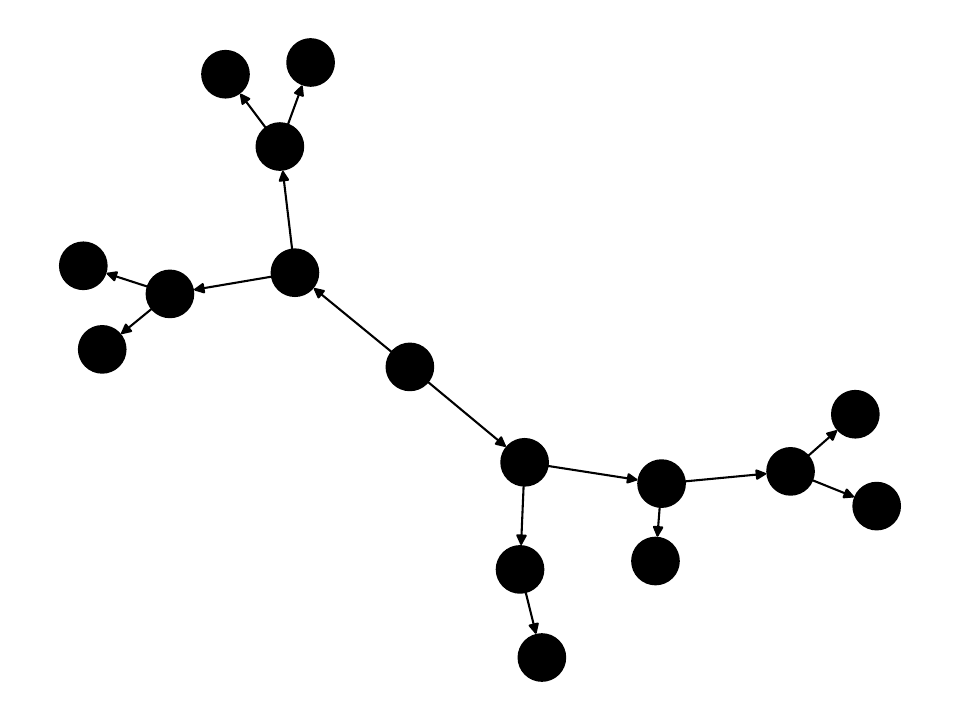}
         \caption{Binary Tree 2}
     \end{subfigure}
     \hfill
     \caption{\textbf{Example graphs from our dataset.}}
     \label{fig:dataset}
\end{figure}

\section{Pooling With Attention}
\label{app:attention}
In addition to the setup described in Section \ref{ssec:pooling}, we also evaluate a simple form of attention where we weight the embeddings produced per node based on a learned function $\phi_\text{att}:\mathbb{R}^{d_\mathbf{h}}\times\mathbb{R}^{d_\text{graphfts}}\rightarrow\mathbb{R}$ depending on both, the node embeddings as well as the encoded graph features.
\begin{equation}
    \mathbf{z}:=\bigoplus_{i\in V}\phi_\text{att}(\mathbf{p}_i^t, \mathbf{h}_g^t)\phi_\text{value}(\mathbf{p}_i^t)
\end{equation}
The underlying idea here is that many algorithms would likely focus on the information of one or few nodes for each step embedding. Apart from that, note that an extensive variety of attention-based mechanisms would be possible. On the one hand, one could change $\phi_\text{att}$ from a simple MLP to a more complex function, e.g. by generating key and value vectors from the node embeddings and query vectors from the graph level features. On the other hand, one could change the representation of graph-level data from the graph features $\mathbf{h}_g^t$ to e.g. some pooled version of the node embeddings $\bigoplus_{i\in V}\mathbf{p}_i^t$ or a combination of these. However, thoroughly evaluating all of these techniques was not the main goal of this work.

In Experiment \ref{row:attention}, we evaluate this attention-based pooling approach. It fails to generalize to larger graphs. Since we use sum pooling, this may be because the magnitude of the encoded graph features are out-of-distribution, making the generated weights perform worse. A mean-pooling approach could lead to better performance.

\section{Implementation Details}
\label{app:impdetails}
Table \ref{tab:imptab} details the values of hyperparameters in our method. Note that complete implementation details are also provided in our repository. The codebase contains a \texttt{README} file with commands to reproduce all of our experiments.

\begin{table}[t]
    \centering
    \caption{\textbf{Parameter values in our network configurations.}}
    \label{tab:imptab}
    \begin{tabular}{cc}
        \toprule
        \textbf{Parameter} & \textbf{Value} \\
        \midrule
        GNN Architecture & MPNN \cite{gilmer2017neural} \\
        Epochs & 20000 \\
        $\bigoplus$ & $\sum$ \\
        $d_{\mathrm{h}}$ & 128 \\
        $d_{\mathrm{stack}}$ & 64 \\
        $\phi_{\mathrm{value}}$ & \{layers: 2,  hidden dim: 128\} \\
        $\phi_{\mathrm{att}}$ & \{layers: 2, hidden dim: 128\} \\
        Activation & ReLU \\
        Optimizer & Adam \citep{DBLP:journals/corr/KingmaB14}  \\
        \bottomrule
    \end{tabular}
\end{table}
\section{\change{Runtime And Memory Consumption}}
\label{app:runtime}
\change{In Table \ref{tab:perf}, we show the runtime and memory consumption of our architecture for all of the ablations we ran in Table \ref{tab:results}. These statistics reveal that using a stack does not increase memory consumption significantly. In fact, our modifications lead to significantly decreased memory consumption when compared to the baseline. This is likely a result of using smaller hints since we converted many node-level hints into graph-level ones. This is also the reason that output collection saves a significant chunk of memory since we do not need to predict the output hints for all nodes.}

\begin{table*}
\centering
 \caption{\change{\textbf{Peak memory consumption (GB) and overall runtime (in minutes) for all training runs. }}}
\label{tab:perf}
\begin{tabular}{l>{(\refstepcounter{rowno}\therowno)}cccccccccc}
\multicolumn{2}{@{}c}{} & \rot{Graph-Level stack (Sec. \ref{ssec:pooling})} & \rot{Node-Wise stack (Sec. \ref{ssec:nodelevel})}  & \rot{Hidden state $\mathbf{p}_i^{t-1}$ (Sec. \ref{ssec:no-recurrent})} & \rot{Output collection (Sec. \ref{ssec:collection})} & \rot{Teacher forcing (50\%)} & \rot{$\phi_\text{value}$ learned (Eq. \ref{eq:no-valuenet})} & \rot{Attention (App. \ref{app:attention})} & \rot{Peak GPU Memory} & \rotatebox{45}{Total Runtime} \\
\toprule
\multirow{1}{*}{\citeauthor{ibarz_generalist_2022}} & \label{row:dfs_orig} & N/A & N/A & \graycheck & \graycross & \graycheck & N/A & N/A & 22.61 & \confint{87.31}{0.12}\\
\hline
\multirow{11}{*}{Ours} & \label{row:ours} & \graycheck & \graycross & \graycross & \graycheck & \graycheck & \graycheck & \graycross & 7.14 & \confint{62.43}{0.10}\\
                      & \label{row:no-stack} & \blackcross & \graycross & \graycross & \graycheck & \graycheck & N/A & N/A & 6.87 & \confint{68.41}{0.09}\\
                      & \label{row:recurrent} & \graycheck & \graycross & \blackcheck & \graycheck & \graycheck & \graycheck & \graycross & 7.56 & \confint{67.36}{0.59}\\
                      & \label{row:recurrent-no-stack} & \blackcross & \graycross & \blackcheck & \graycheck & \graycheck & N/A & N/A & 8.82 & \confint{67.36}{0.59}\\
                      & \label{row:no-collection} & \graycheck & \graycross & \graycross & \blackcross & \graycheck & \graycheck & \graycross & 13.33 & \confint{88.47}{0.10} \\
                      & \label{row:no-tf} & \graycheck & \graycross & \graycross & \graycheck & \blackcross & \graycheck & \graycross & 8.57 & \confint{71.32}{0.16} \\
                      & \label{row:no-valuenet} & \graycheck & \graycross & \graycross & \graycheck & \graycheck & \blackcross & \graycross & 8.45 & \confint{67.82}{0.27}\\
                      & \label{row:attention} & \graycheck & \graycross & \graycross & \graycheck & \graycheck & \graycheck & \blackcheck & 8.75 & \confint{69.68}{0.21}\\
                      & \label{row:nodelevel} & \blackcross & \blackcheck & \graycross & \graycheck & \graycheck & \graycheck & N/A & 7.14 & \confint{79.30}{0.75}\\
                      & \label{row:nodelevel-recurrent} & \blackcross & \blackcheck & \blackcheck & \graycheck & \graycheck & \graycheck & N/A & 8.43 & \confint{80.93}{0.18}\\
\bottomrule
\end{tabular}
\end{table*}

% \begin{table*}[t]
%     \centering
%     \caption{\change{\textbf{Peak memory consumption and overall runtime (in minutes) for all training runs. }}}
%     \begin{tabular}{lccc}
%         \toprule
%          \textbf{Network Configuration} & \parbox[t]{2.5cm}{\textbf{Peak GPU\\ Memory}} & \textbf{Total Runtime}\\
%         \midrule
%         \citet{ibarz_generalist_2022} (single task) & 22.61GB & \confint{87.31}{0.12}\\
%         \hline
%         Ours (baseline) & 7.14GB & \confint{68.41}{0.09}\\
%         Ours $-$ stack & 6.87GB & \confint{62.43}{0.10}\\
%         Ours $+$ hidden state $\mathbf{p}_i^{t-1}$ & 7.56GB & \confint{70.94}{0.18}\\
%         Ours $+$ hidden state $-$ stack & 8.82GB & \confint{67.36}{0.59}\\
%         % \hline
%         Ours $-$ output collection (Section \ref{ssec:collection}) & 13.33GB & \confint{88.47}{0.10} \\
%         Ours $-$ teacher forcing & 8.57GB & \confint{71.32}{0.16} \\
%         Ours \textit{but} $\phi_\text{value}$ not learned (Equation \ref{eq:no-valuenet}) & 8.45GB & \confint{67.82}{0.27}\\
%         Ours $+$ attention (Appendix \ref{app:attention}) & 8.75GB & \confint{69.68}{0.21}\\
%         % \hline
%         Ours $-$ graph-level stack $+$ node-wise stack (Section \ref{ssec:nodelevel}) & 7.14GB & \confint{79.30}{0.75}\\
%         Ours $-$ graph-level stack $+$ node-wise stack $+$ hidden state & 8.43GB & \confint{80.93}{0.18}\\
%         \bottomrule
%     \end{tabular}
%     \label{tab:perf}
% \end{table*}

% \newpage
\section{Depth-First Search Algorithm}
\label{app:dfsalg}

\begin{table}[h]
    % \begin{center}
      % \caption{\textbf{DFS Algorithm.}}
      \renewcommand{\arraystretch}{1.2}
      \label{tab:dfsalg}
      \begin{tabular}{ll}
        % \toprule % <-- Toprule here
        \multicolumn{2}{l}{\textbf{Algorithm 1: Depth-First Search.} Reproduced from \textit{Introduction to algorithms} \citep[Chapter 22]{cormen_introduction_2009}.} \\\\
        % \midrule % <-- Midrule here
        \multicolumn{2}{l}{DFS($G$)} \\
        1 & \textbf{for} each vertex $u \in G.V$ \\
        2 & \hspace{8pt} $u.color = \mathrm{WHITE}$ \\
        3 & \hspace{8pt} $u.\pi = \mathrm{NIL}$ \\
        4 & $time = 0$ \\
        5 & \textbf{for} each vertex $u \in G.V$ \\
        6 & \hspace{8pt} \textbf{if} $u.color == \mathrm{WHITE}$ \\
        7 & \hspace{16pt} DFS-Visit($G$, $u$) \\
        \\
        \multicolumn{2}{l}{DFS-Visit($G$, $u$)} \\
        1 & $time = time + 1$ \\
        2 & $u.d = time$ \\
        3 & $u.color = \mathrm{GRAY}$ \\
        4 & \textbf{for} each $v \in G.Adj[u]$ \\
        5 & \hspace{8pt} \textbf{if} $v.color == \mathrm{WHITE}$ \\
        6 & \hspace{16pt} $v.\pi = u$ \\
        7 & \hspace{16pt} DFS-Visit($G$, $v$) \\
        8 & $u.color = \mathrm{BLACK}$ \\
        9 & $time = time + 1$ \\
        10 & $u.f = time$ \\
        % \bottomrule % <-- Bottomrule here
      \end{tabular}
    % \end{center}
\end{table}

% \clearpage
% \section{\change{Method Pseudocode}}
\definecolor{codegreen}{rgb}{0,0.6,0}
\definecolor{codegray}{rgb}{0.5,0.5,0.5}
\definecolor{codepurple}{rgb}{0.58,0,0.82}
\definecolor{backcolour}{rgb}{0.95,0.95,0.92}

\lstdefinestyle{mystyle}{
    backgroundcolor=\color{backcolour},   
    commentstyle=\color{codegreen},
    keywordstyle=\color{magenta},
    numberstyle=\tiny\color{codegray},
    stringstyle=\color{codepurple},
    basicstyle=\ttfamily\footnotesize,
    breakatwhitespace=false,         
    breaklines=true,                 
    captionpos=t,                    
    keepspaces=true,                 
    numbers=left,                    
    numbersep=5pt,                  
    showspaces=false,                
    showstringspaces=false,
    showtabs=false,                  
    tabsize=2
}

\lstset{style=mystyle}

\begin{lstlisting}[language=Python, float=h, caption=\change{A high-level overview of our method as Python-like pseudocode.}, columns=fullflexible]
result_hint: str
current_node_hint: str
graph_hint_names: List[str]
node_hint_names: List[str]
V: List[Node]
E: List[Tuple[Node, Node]]

graph_hints = {n: None for n in graph_hint_names}
node_hints = {v: {n: None for n in node_hint_names} for v in V}
results = {v: None for v in V}
stack = []
for _ in range(GNN_iter):
    stack_top = stack[-1]
    h_g = sum(f_g[hint_name](hint_value)
              for hint_name, hint_value in graph_hints.items())
    inputs = {}
    for v in V:
        h_i = sum(f_n[hint_name](hint_value)
                  for hint_name, hint_value in node_hints[v].items())
        inputs[v] = concatenate(h_g, h_i, stack_top)
    
    p_i = psi(inputs)
    graph_hints = {k: g(sum(p_i[v] for v in V)) for k, g in g_g.items()}
    node_hints = {k: {v: g(p_i[v]) for v in  V} for k, g in g_n.items()}
    
    if argmax(graph_hints["stack_op"]) == 2: # push
        stack.push(sum(phi_value(p_i)))
    elif argmax(graph_hints["stack_op"]) == 0 and len(stack) > 1: # pop
        stack.pop()
    
    results[argmax(graph_hints[current_node_hint])] = graph_hints[result_hint]
\end{lstlisting}
% def psi(x: dict[Node, vector]):
%     for l in range(num_layers):
%         for v in V:
%             x[v] = g_pooled[l](sum(g_edge[l](x[u], x[v]) for (u, v) in E)
%     return x
\end{document}